\documentclass[aip,cha,reprint,longbibliography,nofootinbib]{revtex4-2}

\usepackage{amsmath,amssymb,bm}
\usepackage{graphicx}
\usepackage{bm}
\usepackage{booktabs}
\usepackage{microtype}
\usepackage{xcolor}
\usepackage{enumitem}
\usepackage[colorlinks=true,citecolor=blue,urlcolor=blue,linkcolor=blue]{hyperref}

\hypersetup{
  colorlinks=true,
  linkcolor=blue!45!black,
  citecolor=blue!45!black,
  urlcolor=blue!55!black,
  pdftitle={NeuroAI Book-- Chapter 5: Memory as an Energy Landscape---Hopfield},
  pdfauthor={Nima Dehghani}
}

\graphicspath{{figures/}}
\newcommand{\dd}{\mathrm{d}}
\newcommand{\Dz}{\mathcal{D}z}
\newcommand{\E}{\mathbb{E}}
\newcommand{\Var}{\mathrm{Var}}

\newcommand{\sgn}{\mathop{\mathrm{sgn}}}
\newcommand{\erf}{\mathop{\mathrm{erf}}}

\newcommand{\softmax}{\mathop{\mathrm{softmax}}}
\newcommand{\order}[1]{\mathcal{O}\!\left(#1\right)}

\begin{document}

\title[\emph{Dehghani, N. -- \textbf{NeuroAI}: Theoretical Foundation of Dynamics, Learning and Computation in Brains, Minds $\&$ Machines}]{Chapter 5: Memory as an Energy Landscape---Hopfield}

\author{Nima Dehghani}
\affiliation{McGovern Institute for Brain Research, Massachusetts Institute of
Technology, Cambridge, Massachusetts 02139, USA}
\affiliation{The NSF AI Institute for Artificial Intelligence and Fundamental
Interactions (IAIFI), Massachusetts Institute of Technology, Cambridge,
Massachusetts 02139, USA}
\email{nima.dehghani@mit.edu}
\date{2026}

\maketitle

\section*{Chapter orientation}

Within \textit{NeuroAI: Theoretical Foundations of Dynamics, Learning and
Computation in Brains, Minds, and Machines}\footnote{For the NeuroAI Book see:\\ \url{https://neurovium.science/books/NeuroAI}}
, this chapter opens the book's
second part, on memory, energy, and collective dynamics. The preceding chapters
trace the formalization of neural computation from threshold logic and the
perceptron to Amari's recurrent dynamics. Hopfield's construction marks the
next conceptual step: a memory becomes a stable collective state of a
many-body system; a corrupted cue becomes an initial condition; recall becomes
a trajectory through state space; and failure becomes a change in basin
geometry. The two chapters that follow use statistical mechanics and dynamic
mean-field theory to ask what happens when disorder, load, and nonequilibrium
dynamics become extensive.

The chapter therefore does more than describe the classical Hopfield network.
It reconstructs the historical problem, derives the binary and graded-response
Lyapunov functions, develops the mean-field account of retrieval and the
$\alpha_c\simeq0.138$ capacity boundary, and follows the energy-based program
through optimization, dense associative memories, exponential interactions,
and modern continuous Hopfield updates. Throughout, capacity claims are tied to
their ensemble and success criterion, and simulations are used to expose
mechanisms rather than substitute for proofs. The aim is to preserve the full
mathematical and physical content while making every assumption and scaling
limit explicit enough to be checked and reproduced.

\section{Memory as a collective dynamical state}
\label{sec:why}

The 2024 Nobel Prize in Physics recognized John J. Hopfield and Geoffrey E.
Hinton ``for foundational discoveries and inventions that enable machine
learning with artificial neural networks.''\cite{Nobel2024} The historical
importance of Hopfield's work is best understood neither by reading
the modern phrase ``Hopfield network'' backward into the 1980s nor by treating
the work as a primitive ancestor of current deep learning. The important change
was conceptual and mathematical: computation became a property of the phase
portrait of a physical system.

Suppose a network has $N$ binary degrees of freedom. Its $2^N$ possible states
are not inspected one by one. The couplings are chosen so that selected global
patterns become attractors. A partial or corrupted cue is an initial condition;
recall is a trajectory; error correction is convergence within a basin; a
memory error is motion into the wrong basin; and capacity is a change in the
typical organization of the landscape as the number of stored patterns grows.
This translation matters. It turns qualitative statements about ``distributed
memory'' into questions with order parameters, stability inequalities,
thermodynamic limits, and phase boundaries.

Hopfield's 1982 PNAS article made this synthesis exceptionally clear for
two-state units, and his 1984 PNAS article showed that the collective mechanism
survives when binary spins are replaced by continuous graded responses.
\cite{Hopfield1982,Hopfield1984} Those short papers contain several different
achievements that are often collapsed into one:

\begin{enumerate}[leftmargin=*]
  \item a dynamical definition of content-addressable memory;
  \item a symmetric recurrent architecture with a Lyapunov function;
  \item a Hebbian construction that embeds memories in the couplings;
  \item a physical interpretation of basins, robustness, and fail-soft behavior;
  \item and, in 1984, a continuous-time circuit model whose energy decreases
  despite graded nonlinear units.
\end{enumerate}

The exact storage limit $0.138N$ was \emph{not} derived in the 1982 paper. It
emerged from the subsequent statistical-mechanical work of Amit, Gutfreund, and
Sompolinsky on an extensive number of random patterns.
\cite{AmitGutfreundSompolinsky1985PRA,AmitGutfreundSompolinsky1985PRL,AmitGutfreundSompolinsky1987}
Similarly, the modern softmax formulation is a later member of the energy-based
family, not an equation found in 1982. Keeping this chronology straight makes
the intellectual development more, not less, impressive.

\begin{figure*}[t]
  \centering
  \includegraphics[width=0.94\textwidth]{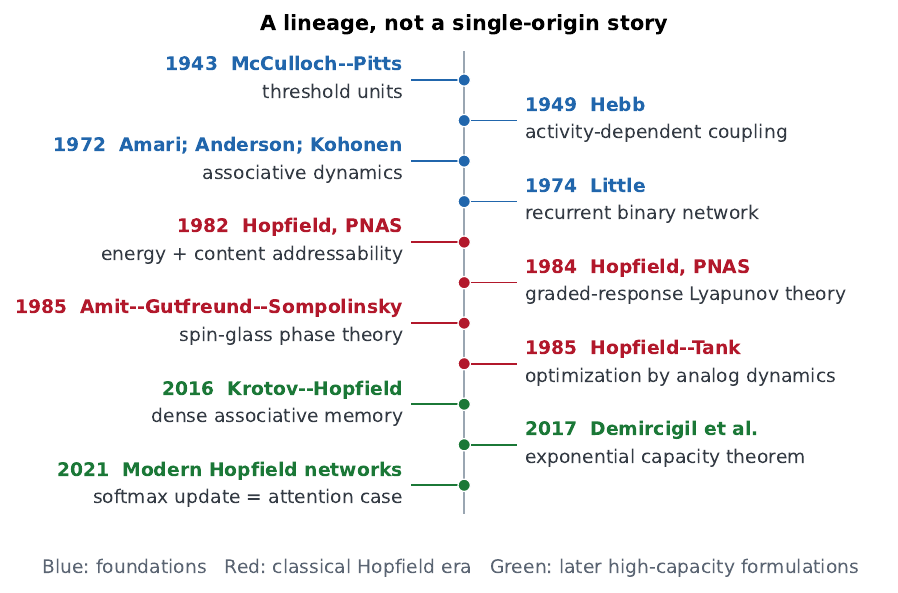}
  \caption{A selective lineage of associative-memory theory. Hopfield did not
  invent threshold neurons, Hebbian association, correlation memories, or the
  first recurrent binary network. The 1982 synthesis made content-addressable
  recall the descent dynamics of a symmetric many-body system and presented its
  collective computational consequences in a form that physicists could
  generalize. The later high-capacity models change the interaction function and
  sometimes the state space, while retaining the idea that memories organize a
  dynamical landscape. This is an original schematic, not data reproduced from
  the cited literature.}
  \label{fig:timeline}
\end{figure*}

\subsection{Prerequisites and route through the chapter}

The mathematical prerequisites are linear algebra, Gaussian random variables,
elementary ordinary differential equations, and the definitions of a partition
function and free energy. No previous study of spin glasses is assumed. A
productive route is:

\begin{enumerate}[leftmargin=*]
  \item derive the one-spin energy change in Sec.~\ref{sec:binary} before
  running the code;
  \item work through the signal--crosstalk decomposition in
  Sec.~\ref{sec:crosstalk}, keeping track of which randomness is quenched;
  \item solve the zero-temperature equations in Sec.~\ref{sec:ags} by
  continuation rather than by independent root finding;
  \item verify the graded-response Lyapunov proof in Sec.~\ref{sec:graded}; and
  \item compare the classical, polynomial dense, and softmax energies in
  Secs.~\ref{sec:dense} and \ref{sec:modern} without assuming that their
  capacity statements use identical criteria.
\end{enumerate}

All numerical panels are generated by
\texttt{code/hopfield\_replication.py}. The figures are pedagogical
replications of model phenomena, not digitizations of published plots.

\section{The historical problem before 1982}
\label{sec:history}

\subsection{Threshold logic, association, and recurrent state}

McCulloch and Pitts formalized threshold elements in 1943, while Hebb's 1949
proposal connected joint activity to persistent changes of coupling.
\cite{McCullochPitts1943,Hebb1949} By the early 1970s, Amari, Anderson, and
Kohonen had developed important recurrent or correlation-based associative
memories, and Little had studied persistent states in a recurrent binary neural
network.\cite{Amari1972,Anderson1972,Kohonen1972,Little1974} It would therefore
be historically wrong to say that associative memory, recurrent networks, or
symmetric interactions began with Hopfield.

What the Hopfield formulation accomplished was a particular unification. The
network was simultaneously:

\begin{itemize}[leftmargin=*]
  \item a neural model, because its couplings could be written as a sum of
  correlations between activity patterns;
  \item a physical system, because symmetric couplings supplied an energy and
  asynchronous updates supplied dissipative motion;
  \item a memory, because selected patterns were fixed points with basins; and
  \item a computational device, because a basin performed error correction and
  content-addressable completion in parallel.
\end{itemize}

The importance of the energy is not that every neural system must minimize one.
It is that a global statement---convergence---can be proved without solving all
$N$ coupled trajectories. That proof identifies which assumptions do the work:
symmetry, the absence of self-coupling, and an update rule consistent with the
local field.

\subsection{Hopfield's style of physical reasoning}

The neural papers also fit a longer scientific trajectory. Hopfield's earlier
work ranged from collective excitations in solids to electron transfer and
kinetic proofreading in molecular biology.\cite{Hopfield1958,HopfieldElectronTransfer1974,HopfieldProofreading1974}
Across these topics, microscopic events are organized by a physical constraint
into a robust macroscopic function: an optical response, a transfer rate, an
error-suppression mechanism, or a memory attractor. The neural-network work did
not merely borrow terminology from magnetism. It asked which physical structure
makes a computation reliable despite many interacting parts.

Hopfield later described productive work at the physics--biology interface as
movement between two scientific cultures: one seeks economical principles and
the other insists on the historically contingent detail of living systems.
His Nobel lecture makes the complementary methodological point that physics is
not only a catalogue of subject matter but a way of choosing variables and
asking which collective regularities survive microscopic complexity.
Bialek's account of the 2024 prize places the neural-network work in precisely
this tradition of emergence at intermediate scales.
\cite{Hopfield2014TwoCultures,Hopfield2025NobelLecture,Bialek2025Emergence}
This perspective helps explain both the power and the restraint of the model:
it searches for a minimal sufficient mechanism, while leaving open whether a
particular neural circuit realizes that mechanism at the chosen scale.

This also explains the repeated emphasis on ``emergent collective
computational abilities'' in the 1982 title. A single model neuron does not
possess a memory basin. A basin is a property of the coupled system. Damage to a
few units or corruption of a few bits can leave the basin intact, which is the
physical source of distributed, fail-soft performance.

\section{The 1982 PNAS model: binary memory as energy descent}
\label{sec:binary}

\subsection{Centered variables and notation}

Hopfield's paper used two-state variables closely related to firing/nonfiring
values $V_i\in\{0,1\}$. For the statistical mechanics it is convenient to use
centered spins
\begin{equation}
  s_i=2V_i-1\in\{-1,+1\},\qquad i=1,\ldots,N.
  \label{eq:centered}
\end{equation}
Let $J_{ij}$ be the influence of unit $j$ on unit $i$, and let $\theta_i$ be a
threshold. We impose
\begin{equation}
  J_{ij}=J_{ji},\qquad J_{ii}=0.
  \label{eq:symmetry}
\end{equation}
The local field is
\begin{equation}
  h_i(\bm{s})=\sum_{j\ne i}J_{ij}s_j-\theta_i.
  \label{eq:localfield}
\end{equation}
At zero temperature, an asynchronous update chooses one index $i$ and sets
\begin{equation}
  s_i\leftarrow \sgn h_i,
  \label{eq:async}
\end{equation}
with any fixed convention when $h_i=0$. ``Asynchronous'' means that the field
used for the next update includes the changes already made. It does not require
a single biological neuron to own a global clock.

The associated energy is
\begin{equation}
  E(\bm{s})=-\frac{1}{2}\sum_{i\ne j}J_{ij}s_i s_j
  +\sum_i\theta_i s_i.
  \label{eq:classicalenergy}
\end{equation}
The factor $1/2$ prevents double counting. Only relative energies matter; adding
a constant changes no dynamics.

\subsection{The Lyapunov proof, one spin at a time}

Suppose only $s_i$ changes, from $s_i$ to $s_i'$. Symmetry lets all terms
involving $i$ be collected:
\begin{align}
  \Delta E
  &=E(s_i',\bm{s}_{\setminus i})-E(s_i,\bm{s}_{\setminus i})\\
  &=-(s_i'-s_i)\left(\sum_{j\ne i}J_{ij}s_j-\theta_i\right)\\
  &=-(s_i'-s_i)h_i.
  \label{eq:deltaenergygeneral}
\end{align}
If the update flips a spin, $s_i'=-s_i$, so
\begin{equation}
  \Delta E=2s_i h_i<0
  \label{eq:deltaenergyflip}
\end{equation}
whenever $s_i$ disagrees with the nonzero field. If it already agrees, no update
is made. Thus $E$ cannot increase. Because there are finitely many states, the
asynchronous deterministic dynamics must reach a fixed point after finitely
many energy-decreasing flips, apart from neutral moves that can be eliminated by
the tie convention.

The proof has a sharply delimited scope:

\begin{enumerate}[leftmargin=*]
  \item Symmetry is used in Eq.~\eqref{eq:deltaenergygeneral}. With asymmetric
  $J$, the same scalar energy does not generally exist.
  \item The proof is for sequential updates. Fully synchronous updates can
  produce two-cycles even when $J$ is symmetric.
  \item Convergence means arrival at a local minimum, not necessarily the
  desired memory or the global minimum.
\end{enumerate}

Together these conditions separate an equilibrium attractor network from
generic recurrent computation.

\paragraph{Scope checkpoint: model and brain.}
Already at this stage it is useful to answer the question ``is this the
brain?'' The binary network is an effective circuit theorem, not a microscopic
reconstruction of cortex. Its biological content is the testable possibility
that a population may possess collective basins and error-correcting relaxation
even when individual neurons are noisy and heterogeneous. Exact binary states,
pairwise symmetry, and asynchronous descent are sufficient conditions in the
model; they are not asserted to be literal properties of every memory circuit.
Section~\ref{sec:limits} returns to the scale at which such an attractor
description can remain closed.

\subsection{Writing memories into the couplings}

Let $p$ desired memories be random or structured binary vectors
\begin{equation}
  \bm{\xi}^{\mu}=(\xi_1^\mu,\ldots,\xi_N^\mu),
  \qquad \xi_i^\mu\in\{-1,+1\},
  \qquad \mu=1,\ldots,p.
\end{equation}
The Hebbian outer-product prescription is
\begin{equation}
  J_{ij}=\frac{1}{N}\sum_{\mu=1}^{p}\xi_i^\mu\xi_j^\mu,
  \qquad i\ne j,\qquad J_{ii}=0.
  \label{eq:hebb}
\end{equation}
The normalization makes the field order one when $p$ grows proportionally to
$N$. Substitution into Eq.~\eqref{eq:classicalenergy}, ignoring the diagonal
constant, gives
\begin{equation}
  E(\bm{s})=-\frac{N}{2}\sum_{\mu=1}^{p}(m^\mu)^2,
  \qquad
  m^\mu=\frac{1}{N}\sum_{i=1}^{N}\xi_i^\mu s_i.
  \label{eq:energyoverlaps}
\end{equation}
The overlap $m^\mu$ is an order parameter. It is $1$ at memory $\mu$, near zero
for an unrelated random state, and intermediate inside a retrieval basin.
Equation~\eqref{eq:energyoverlaps} makes the design transparent: lowering energy
rewards large squared overlap with one or more stored patterns.

The phrase ``the memory is stored in every synapse'' should be used carefully.
Each $J_{ij}$ is a superposition of contributions from all patterns. No single
coupling contains a readable copy. The memory is expressed collectively in the
field generated when many current-state bits align with the same pattern.

\begin{figure*}[t]
  \centering
  \includegraphics[width=0.94\textwidth]{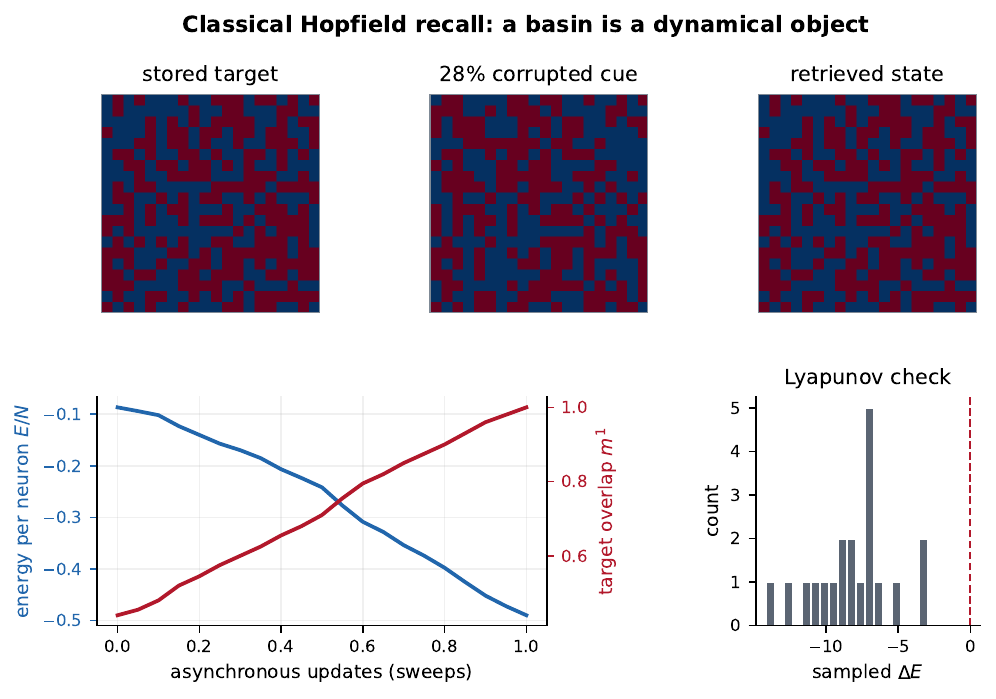}
  \caption{Seeded replication of classical content-addressable recall. Eighteen
  random patterns are embedded in $N=400$ spins. A cue obtained by flipping
  $28\%$ of one memory begins with overlap $m^1=0.44$ and converges to the stored
  target under random sequential updates. The energy decreases at every sampled
  interval while the overlap grows; in this run the final overlap is $1$. A
  single successful cue demonstrates a basin, not a storage-capacity theorem.
  Source: \texttt{code/hopfield\_replication.py}.}
  \label{fig:classical}
\end{figure*}

\section{Signal, crosstalk, basins, and false memories}
\label{sec:crosstalk}

\subsection{The local field near a stored pattern}

To test retrieval of pattern $\mu=1$, insert Eq.~\eqref{eq:hebb} into the field
with zero thresholds:
\begin{align}
  h_i
  &=\sum_{j\ne i}\frac{1}{N}\sum_{\mu=1}^{p}
  \xi_i^\mu\xi_j^\mu s_j\\
  &\simeq \xi_i^1 m^1
  +\sum_{\mu=2}^{p}\xi_i^\mu m^\mu.
  \label{eq:signalnoise}
\end{align}
The first term is the retrieval signal. The remaining memories produce
crosstalk. If $\bm{s}=\bm{\xi}^1$ and all patterns are independent unbiased
Rademacher variables, then for $\mu>1$
\begin{equation}
  m^\mu=\frac{1}{N}\sum_j\xi_j^\mu\xi_j^1
\end{equation}
has mean zero and variance $1/N$. For $p=\alpha N$, the sum of roughly $p$
small terms in Eq.~\eqref{eq:signalnoise} has order-one variance
\begin{equation}
  \Var(\text{crosstalk})\simeq\frac{p}{N}=\alpha.
  \label{eq:naivevariance}
\end{equation}
The central limit theorem suggests the local stability estimate
\begin{equation}
  \Pr(\xi_i^1 h_i<0)\approx
  \Phi\!\left(-\frac{1}{\sqrt{\alpha}}\right),
  \label{eq:naiveerror}
\end{equation}
where $\Phi$ is the standard normal cumulative distribution.
The random variable $\xi_i^1h_i$ is the signed stability margin of bit $i$:
positive values support the target bit and negative values flip it. In the
naive Gaussian picture its mean is one and its variance is $\alpha$, which is
why Eq.~\eqref{eq:naiveerror} has the displayed signal-to-noise ratio.

This calculation is indispensable, but it does not by itself yield
$\alpha_c=0.138$. It treats crosstalk terms too independently and asks about
local stability at a prescribed configuration. During retrieval the state and
the disorder become correlated; the network's susceptibility feeds errors back
into the noise. The full mean-field theory introduces an effective noise
variance $\alpha r$, not simply $\alpha$.

\subsection{A basin is not a number attached to a pattern}

The basin of memory $\mu$ is the set
\begin{equation}
  \mathcal{B}_\mu=\left\{\bm{s}_0:\ \lim_{t\to\infty}\bm{s}(t)=\bm{s}_\mu^*\right\},
  \label{eq:basin}
\end{equation}
where $\bm{s}_\mu^*$ is the attractor associated with that memory. Basin size
depends on the update rule, temperature, thresholds, pattern correlations,
coding level, and the measure used to draw cues. Hamming radius is convenient
for random binary patterns, but it is not universally the natural geometry.

Several commonly quoted ``capacities'' therefore answer different questions:

\begin{itemize}[leftmargin=*]
  \item Are the nominal patterns exact fixed points?
  \item Is every bit of every pattern stable with high probability?
  \item Does a typical corrupted cue recover a state with macroscopic overlap?
  \item Is the retrieval state thermodynamically dominant?
  \item How much mutual information can be recovered?
\end{itemize}

A scaling law without its success criterion is incomplete.

\subsection{Spurious minima and mixture states}

Equation~\eqref{eq:energyoverlaps} rewards squared overlaps. It does not demand
that only one overlap be nonzero. For an odd set of nearly orthogonal random
patterns, a majority-vote mixture
\begin{equation}
  s_i^{(123)}=\sgn(\xi_i^1+\xi_i^2+\xi_i^3)
  \label{eq:mixture}
\end{equation}
has, in the large-$N$ limit, overlap $m^1=m^2=m^3=1/2$. The corresponding
leading energy is approximately $-3N/8$, compared with $-N/2$ for a pure
memory. Such states can nevertheless be locally stable. There are also
spin-glass-like minima with no macroscopic overlap with any one pattern.

This is why ``energy minima are memories'' is too strong. The designed memories
are intended minima; the nonlinear superposition creates additional minima.
Above useful retrieval capacity, the landscape does not become flat. It remains
rugged, but the minima cease to support reliable identification with the stored
patterns.

\section{Temperature, quenched disorder, and the meaning of 0.138}
\label{sec:ags}

\subsection{From deterministic descent to an equilibrium ensemble}

A stochastic single-spin update can be chosen as
\begin{equation}
  \Pr(s_i=\pm1\mid\bm{s}_{\setminus i})
  =\frac{e^{\pm\beta h_i}}{2\cosh(\beta h_i)},
  \label{eq:glauber}
\end{equation}
which obeys detailed balance with
\begin{equation}
  P(\bm{s})=\frac{1}{Z}e^{-\beta E(\bm{s})},
  \qquad
  Z=\sum_{\bm{s}}e^{-\beta E(\bm{s})}.
  \label{eq:boltzmann}
\end{equation}
Here $T=\beta^{-1}$ is literal model temperature: it controls stochastic state
updates. It should not automatically be identified with biological temperature,
observation noise, learning rate, or minibatch noise.

For a finite number of patterns, a one-condensed-pattern mean-field argument
gives the Curie--Weiss-like equation
\begin{equation}
  m=\tanh(\beta m).
  \label{eq:lowloadmf}
\end{equation}
The nonzero solution appears below $T=1$ in the chosen units. But the technically
interesting regime is $p=\alpha N$, where the disorder remains extensive.

The patterns are \emph{quenched}: one draws them, constructs $J$, and studies
the network for that fixed realization. The typical free energy is therefore
\begin{equation}
  f=-\frac{1}{\beta N}\E_{\xi}[\ln Z(\xi)],
  \label{eq:quenchedfree}
\end{equation}
not $-(\beta N)^{-1}\ln\E_\xi[Z]$. The logarithm prevents a direct disorder
average. The replica identity
\begin{equation}
  \E[\ln Z]=\lim_{n\to0}\frac{\E[Z^n]-1}{n}
  \label{eq:replica}
\end{equation}
turns the problem into coupled copies, after which overlaps between replicas
emerge as saddle-point variables. The method does not average the disorder
away; it converts its typical effects into order parameters.

\subsection{Replica-symmetric retrieval equations}

\begin{center}
\setlength{\fboxsep}{5pt}
\fcolorbox{blue!35!black}{blue!3}{%
\parbox{0.90\columnwidth}{\small
\textbf{Reader's map: how the replica reduction reaches one effective neuron.}
For integer $n$, $Z^n$ introduces replicas $a=1,\ldots,n$. One first singles
out a condensed memory through $m_a=N^{-1}\sum_i\xi_i^1s_i^a$ and records the
mutual overlaps $q_{ab}=N^{-1}\sum_i s_i^as_i^b$. Averaging the remaining
$\alpha N-1$ random patterns couples the replicas; auxiliary Gaussian fields
then decouple the sites. The $N$-spin trace factorizes into copies of one
effective neuron driven by a deterministic retrieval field $m$ plus a Gaussian
crosstalk field $z\sqrt{\alpha r}$. Replica symmetry sets $m_a=m$ and
$q_{a\ne b}=q$. Stationarity of the resulting free energy gives
Eqs.~\eqref{eq:rsm}--\eqref{eq:rsr}. Thus $r$ is not an extra phenomenological
noise parameter: it is the variance generated after recurrent susceptibility
dresses the bare crosstalk. The full disorder average and saddle-point algebra
are developed in Chapter 6, \emph{Sompolinsky: Associative-Memory Statistical
Mechanics}; the present box supplies the logical bridge to the equations used
here.}}
\end{center}

For one condensed memory, the replica-symmetric equations can be written
\begin{align}
  m&=\int\Dz\,\tanh\!\left[\beta(m+z\sqrt{\alpha r})\right],
  \label{eq:rsm}\\
  q&=\int\Dz\,\tanh^2\!\left[\beta(m+z\sqrt{\alpha r})\right],
  \label{eq:rsq}\\
  C&=\beta(1-q),
  \label{eq:rsc}\\
  r&=\frac{q}{(1-C)^2},
  \label{eq:rsr}
\end{align}
where
\begin{equation}
  \Dz=\frac{e^{-z^2/2}}{\sqrt{2\pi}}\dd z.
\end{equation}
The variables have distinct meanings. $m$ is retrieval overlap; $q$ measures
frozen local polarization; $C$ is an integrated single-site susceptibility;
and $r$ renormalizes the variance of the crosstalk field. The denominator in
Eq.~\eqref{eq:rsr} is the feedback missing from the elementary estimate
Eq.~\eqref{eq:naivevariance}.
The first two equations are averages over that effective neuron: the same
Gaussian field $z$ determines its mean response and its squared response.
Equations~\eqref{eq:rsc} and \eqref{eq:rsr} then close the feedback loop by
turning local susceptibility into a dressed noise variance.

As $T\to0$, $q\to1$ but $C=\beta(1-q)$ remains finite. The equations reduce to
\begin{align}
  m&=\erf\!\left(\frac{m}{\sqrt{2\alpha r}}\right),
  \label{eq:t0m}\\
  C&=\sqrt{\frac{2}{\pi\alpha r}}
  \exp\!\left(-\frac{m^2}{2\alpha r}\right),
  \label{eq:t0c}\\
  r&=\frac{1}{(1-C)^2}.
  \label{eq:t0r}
\end{align}
In this limit $\tanh(\beta h)$ becomes $\sgn h$. The Gaussian average of that
threshold produces the error function in Eq.~\eqref{eq:t0m}; the exponentially
small density of sites near zero field leaves the finite susceptibility in
Eq.~\eqref{eq:t0c}. The apparent three-variable system is therefore the
zero-temperature closure of one signal field, one dressed noise scale, and the
response that couples them.
Following the nonzero solution by continuation in $\alpha$ gives a terminal
point near
\begin{equation}
  \alpha_c\simeq0.138.
  \label{eq:alphac}
\end{equation}
This is the celebrated zero-temperature dynamical retrieval spinodal for dense,
symmetric, Hebbian storage of independent unbiased random patterns in the
thermodynamic limit. It is not a universal constant of associative memory.
Different learning rules can approach $p=N$ for appropriate criteria;
sparse coding and structured patterns change the scaling or coefficient; and
thermodynamic dominance of retrieval can be lost before the metastable retrieval
branch disappears.\cite{Personnaz1986,Gardner1988,TsodyksFeigelman1988,McEliece1987}

\begin{figure*}[t]
  \centering
  \includegraphics[width=0.94\textwidth]{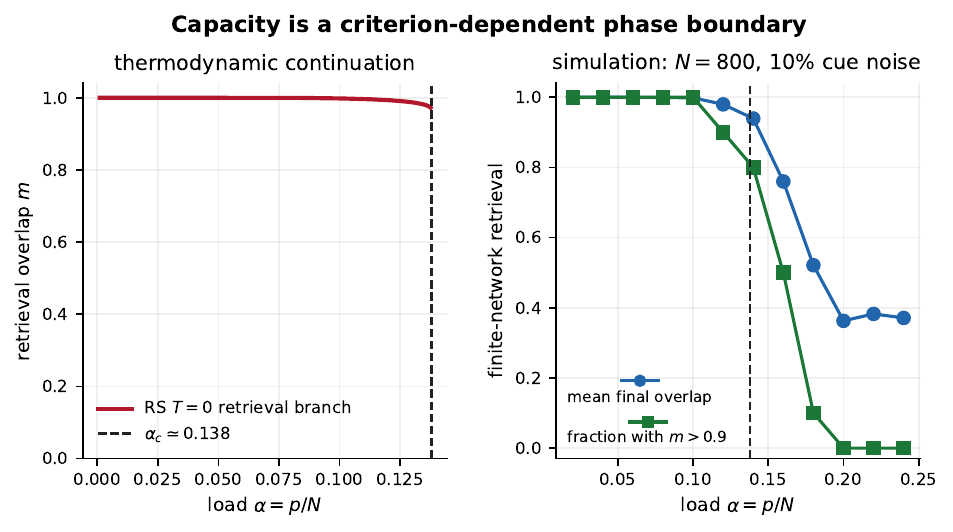}
  \caption{Two complementary replications of capacity. Left: numerical
  continuation of Eqs.~\eqref{eq:t0m}--\eqref{eq:t0r}; the nonzero retrieval
  branch terminates at $\alpha\approx0.1376$ at the plotted resolution. Right:
  finite-$N$ synchronous simulations from cues with $10\%$ flipped bits. The
  finite network rounds and shifts the transition, and the mean-overlap and
  $m>0.9$ criteria do not coincide. The dashed line is a thermodynamic reference,
  not a fit to finite-$N$ points. Source:
  \texttt{code/hopfield\_replication.py}.}
  \label{fig:capacity}
\end{figure*}

\subsection{What the phase language adds}

The useful macroscopic regimes are distinguished by order parameters:

\begin{center}
\begin{tabular}{lcc}
\toprule
Regime & $m$ & $q$ \\
\midrule
Paramagnetic & $0$ & $0$ \\
Retrieval & $\ne0$ & $>0$ \\
Spin glass & $0$ & $>0$ \\
\bottomrule
\end{tabular}
\end{center}

This is more precise than drawing a landscape with a few wells. A retrieval
phase exists when one memory has order-one overlap in the limit $N\to\infty$.
A spin-glass phase is frozen but not aligned with a designated memory. The two
can have many microscopic minima while differing sharply in computational
meaning.

Replica symmetry is an ansatz, not an identity. In the glassy regime it can be
unstable, and replica-symmetry breaking may be needed to describe the hierarchy
of pure states. The retrieval spinodal remains a useful landmark, but the full
low-temperature glass structure is richer than the three scalar equations
suggest. Quantitatively, however, the correction to this particular capacity
landmark is small: a corrected one-step RSB calculation gives
$\alpha_c\simeq0.138186$ (two-step RSB gives $0.138187$), compared with the
replica-symmetric $0.137905$; finite-size simulations are consequently often
summarized as a capacity near $0.14$.\cite{Amit1989,SteffanKuhn1994}

\section{The 1984 PNAS paper: graded responses and circuit dynamics}
\label{sec:graded}

\subsection{Continuous neurons}

The 1982 construction might have been dismissed as dependent on idealized
binary elements. Hopfield's 1984 paper addressed that objection directly. Let
$u_i$ be an internal potential and
\begin{equation}
  V_i=g_i(u_i)
  \label{eq:gradedoutput}
\end{equation}
be a monotone graded output, such as a sigmoid. A circuit equation is
\begin{equation}
  C_i\frac{\dd u_i}{\dd t}
  =\sum_j T_{ij}V_j-\frac{u_i}{R_i}+I_i,
  \label{eq:gradedode}
\end{equation}
where $C_i$ and $R_i$ can be read as capacitance and leak resistance, $T_{ij}$
as symmetric transconductance, and $I_i$ as external input. The threshold-like
nonlinearity now lies in $g_i$ rather than in an instantaneous sign update.

For $T_{ij}=T_{ji}$ and monotone invertible $g_i$, define
\begin{equation}
  E(\bm{V})=
  -\frac{1}{2}\sum_{ij}T_{ij}V_iV_j
  +\sum_i\frac{1}{R_i}\int_0^{V_i}g_i^{-1}(v)\dd v
  -\sum_i I_iV_i.
  \label{eq:gradedenergy}
\end{equation}
The three terms have direct roles: recurrent interaction lowers the first,
the integral of the inverse response supplies the single-unit leak cost, and
external current tilts the landscape. The inverse-response term is the crucial
construction: differentiating it with respect to $V_i$ returns
$g_i^{-1}(V_i)=u_i$, exactly the variable that appears in the leak.
Differentiation gives
\begin{equation}
  \frac{\partial E}{\partial V_i}
  =-\sum_jT_{ij}V_j+\frac{u_i}{R_i}-I_i
  =-C_i\dot u_i.
  \label{eq:gradedgradient}
\end{equation}
Since $\dot V_i=g_i'(u_i)\dot u_i$,
\begin{equation}
  \frac{\dd E}{\dd t}
  =-\sum_i C_i g_i'(u_i)(\dot u_i)^2\le0.
  \label{eq:gradeddescent}
\end{equation}
The conclusion is structural: the energy argument does not require binary
states. It requires symmetric coupling and monotone input--output functions.
Cohen and Grossberg developed related general stability results for continuous
competitive networks.\cite{CohenGrossberg1983}

\begin{figure*}[t]
  \centering
  \includegraphics[width=0.94\textwidth]{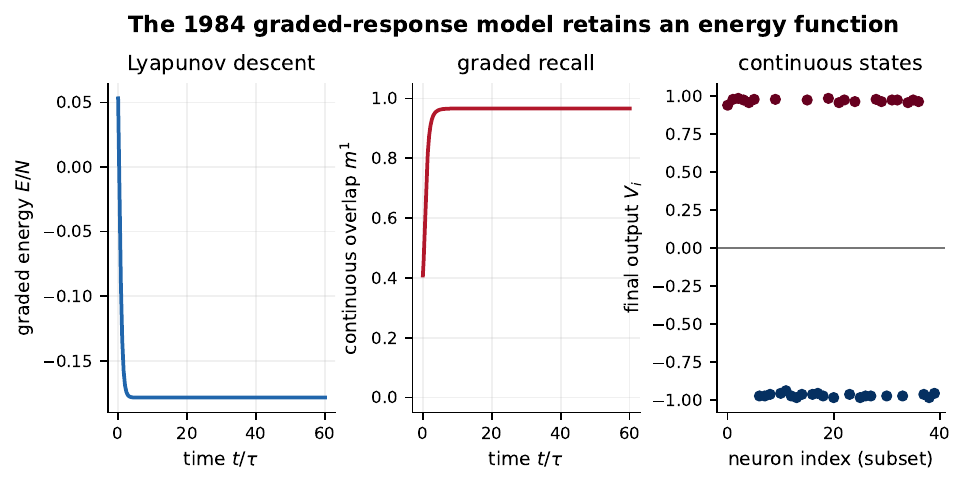}
  \caption{Replication of the graded-response Lyapunov mechanism. Four random
  patterns are embedded in $N=120$ units with $g(u)=\tanh(2.2u)$. From a
  corrupted continuous cue, the energy in Eq.~\eqref{eq:gradedenergy} decreases
  to numerical precision and the target overlap approaches $0.966$. The units
  remain continuous even though the attractor lies close to saturated values.
  This figure tests the 1984 structural claim, not a biological fit. Source:
  \texttt{code/hopfield\_replication.py}.}
  \label{fig:graded}
\end{figure*}

For the simulation in Fig.~\ref{fig:graded}, $R_i=C_i=1$ and
$g(u)=\tanh(\gamma u)$. The integral is explicit:
\begin{equation}
  \int_0^V g^{-1}(v)\dd v
  =\frac{1}{\gamma}
  \left[V\,\mathop{\mathrm{atanh}}V+\frac{1}{2}\ln(1-V^2)\right].
  \label{eq:tanhintegral}
\end{equation}

\subsection{What generalizes, and what remains idealized}

The graded model strengthens the collective argument: attractor computation is
not an artifact of exact McCulloch--Pitts states. But continuous does not mean
biophysically complete. The units remain single variables with static monotone
input--output curves; the couplings are symmetric; delays and adaptation are
absent; and the energy excludes generic oscillations or chaos. The model is an
effective theory of a computational regime, not a claim that a cortical circuit
must literally implement Eq.~\eqref{eq:gradedode} at every microscopic scale.

This distinction is scientifically productive. The right question is not
whether a real neuron is ``a Hopfield neuron.'' It is whether the circuit and
timescale of interest support slow collective variables, recurrent feedback,
and attractors that can be operationally tested by perturbing initial
conditions. More detailed cellular models can change basin geometry or destroy
the Lyapunov structure; that is a prediction about scale, not a failure of
abstraction.

\section{From memory to optimization}
\label{sec:optimization}

Hopfield and Tank extended the energy idea to analog optimization.
\cite{HopfieldTank1985} If a combinatorial problem has binary decision variables
$x_i$ and a cost
\begin{equation}
  \mathcal{C}(\bm{x})=
  -\frac{1}{2}\sum_{ij}W_{ij}x_ix_j+\sum_i b_i x_i
  +\sum_a\lambda_a\,\mathcal{P}_a(\bm{x}),
  \label{eq:optimizationcost}
\end{equation}
one can choose network couplings so that the neural energy represents the
objective plus penalties $\mathcal{P}_a$ for violated constraints. The network
then performs parallel local search by physical relaxation.

The traveling-salesperson construction illustrates both the appeal and the
difficulty. A binary matrix indicates whether city $X$ occupies tour position
$i$. Energy terms penalize more than one city per position, more than one
position per city, and the wrong total number of active units, while a distance
term rewards short consecutive edges. The network does not ``understand'' a
tour. Feasible tours are encoded as low-energy corners of a larger continuous
state space.

Three lessons survive in current energy-based machine learning.

\begin{enumerate}[leftmargin=*]
  \item \emph{Representation is part of the algorithm.} A poor encoding creates
  bad minima even if the dynamics descends perfectly.
  \item \emph{Constraint scales matter.} If penalties are too weak, invalid
  states win; if too strong, the landscape can become stiff and hard to explore.
  \item \emph{Descent is not global optimization.} A Lyapunov proof guarantees
  convergence to a stationary point, not the optimum of an NP-hard problem.
\end{enumerate}

The optimization work was therefore historically important without supplying a
general polynomial-time solver. Its lasting contribution is the explicit
translation between a computational objective and a physical landscape.

\section{Higher-order interactions and dense associative memory}
\label{sec:dense}

\subsection{Changing the shape of a memory well}

The classical energy can be written as a sum of a quadratic function of each
memory overlap. Krotov and Hopfield generalized it to
\begin{equation}
  E_F(\bm{\sigma})=-\sum_{\mu=1}^{K}
  F\!\left(\sum_{i=1}^{N}\xi_i^\mu\sigma_i\right),
  \label{eq:denseenergy}
\end{equation}
where $\sigma_i\in\{-1,+1\}$ and $F$ grows more sharply than a quadratic.
\cite{KrotovHopfield2016} For $F(x)=x^n$, the energy is a degree-$n$ polynomial
in the spins. The $n=2$ case reduces, up to scale and diagonal constants, to the
classical Hebbian model. Increasing $n$ makes a strongly matching memory
dominate over many weakly matching memories.

The energy-correct asynchronous update must compare the two possible energies
of spin $i$. Define
\begin{equation}
  S_{\mu i}=\sum_{j\ne i}\xi_j^\mu\sigma_j.
\end{equation}
Then
\begin{equation}
  \sigma_i\leftarrow\sgn\sum_{\mu=1}^{K}
  \left[F(S_{\mu i}+\xi_i^\mu)-F(S_{\mu i}-\xi_i^\mu)\right].
  \label{eq:denseupdate}
\end{equation}
This is an exact energy-difference rule. Replacing it by a formal derivative of
$F$ can change self-interaction terms at finite $N$.

\subsection{Why the capacity exponent changes}

Initialize the network exactly at memory $\nu$. For one bit, the target memory
contributes an energy gap
\begin{equation}
  \E[\Delta E_{\mathrm{signal}}]
  =N^n-(N-2)^n
  \simeq 2nN^{n-1}.
  \label{eq:densesignal}
\end{equation}
For a non-target random memory, its overlap with the target is approximately a
Gaussian $z\sim\mathcal{N}(0,N)$. Expanding the difference of powers gives the
leading fluctuation $2n\xi_i z^{n-1}$. Since
\begin{equation}
  \E[z^{2n-2}]=(2n-3)!!\,N^{n-1},
\end{equation}
the crosstalk variance from $K$ patterns scales as
\begin{equation}
  \Var(\Delta E_{\mathrm{noise}})
  \simeq4n^2K(2n-3)!!\,N^{n-1}.
  \label{eq:densevariance}
\end{equation}
The signal-to-noise ratio is therefore
\begin{equation}
  \mathrm{SNR}\sim
  \sqrt{\frac{N^{n-1}}{K(2n-3)!!}}.
  \label{eq:densesnr}
\end{equation}
Keeping a fixed bit-error probability permits
\begin{equation}
  K_{\max}=\alpha_n N^{n-1},
  \label{eq:densefixed}
\end{equation}
where the constant depends on the allowed error. Requiring an entire pattern to
have no errors with high probability introduces an extreme-value correction:
\begin{equation}
  K_{\max}^{\mathrm{no\ error}}\simeq
  \frac{1}{2(2n-3)!!}\frac{N^{n-1}}{\ln N}.
  \label{eq:densenone}
\end{equation}
The logarithm is the price of changing the event being controlled. A fixed
bit only requires its Gaussian tail probability to be small; demanding that
none of $N$ bits fail requires a tail of order $1/N$ (by a union or
extreme-value estimate), which costs the extra $\ln N$.
Thus a cubic energy can have order $N^2$ fixed-error capacity and a quartic
energy order $N^3$, rather than order $N$. The exponent is the central result;
the prefactor depends on energy normalization, update convention, correlations,
and the retrieval criterion. Earlier higher-order memory models anticipated
parts of this scaling family; the dense-associative-memory work connected it
directly to trainable pattern recognition and modern neural architectures.
\cite{Kanter1988,KrotovHopfield2016}

\begin{figure*}[t]
  \centering
  \includegraphics[width=0.94\textwidth]{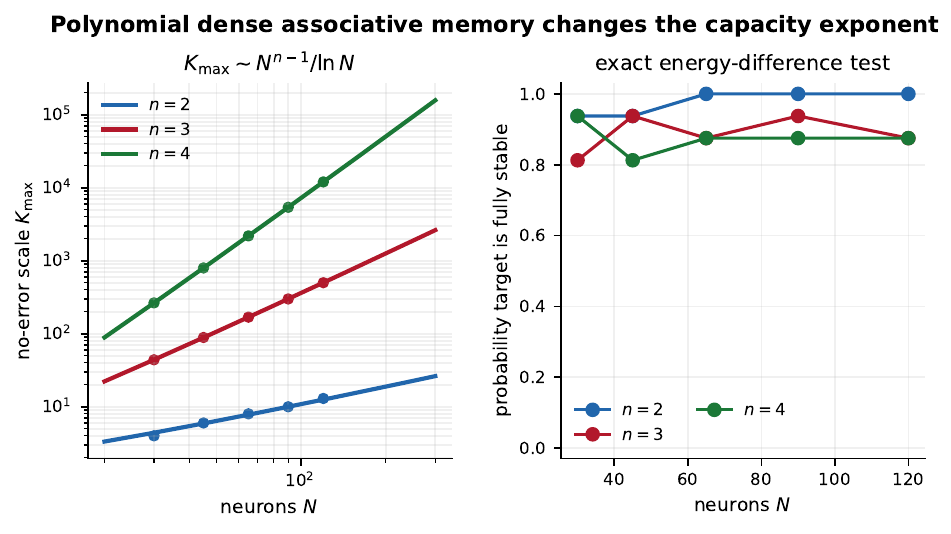}
  \caption{Polynomial dense-memory capacity. Left: the no-error scale in
  Eq.~\eqref{eq:densenone} for interaction orders $n=2,3,4$. Right: direct
  finite-$N$ tests of whether every bit of a designated random target is stable
  under the exact energy-difference rule Eq.~\eqref{eq:denseupdate}, with $K$
  set to the asymptotic no-error scale. The finite-size probabilities are noisy
  and the asymptotic coefficient is conservative at these sizes; the robust
  claim is the changed power of $N$. Source:
  \texttt{code/hopfield\_replication.py}.}
  \label{fig:dense}
\end{figure*}

\subsection{From features to prototypes}

Krotov and Hopfield also showed a dual description of one-step dense-memory
classification as a feedforward network with a hidden layer.
\cite{KrotovHopfield2016} The derivative or finite difference of $F$ becomes the
hidden-unit activation. For rectified polynomial $F$, small powers let several
stored directions cooperate like distributed features; at large powers the
largest-overlap term dominates, producing a prototype-like computation. This is
not merely a capacity increase. Changing $F$ changes the qualitative algorithm
implemented by the landscape.

The same perspective motivated work on adversarial inputs: high-order energies
can shape low-energy configurations more tightly around meaningful training
patterns.\cite{KrotovHopfield2018} The conclusion should remain model-specific.
Robustness in a dense associative memory does not imply that arbitrary networks
with high-degree activations are automatically robust.

\subsection{The many-body synapse problem}

Expanding Eq.~\eqref{eq:denseenergy} for $F(x)=x^n$ produces tensors such as
\begin{equation}
  T_{i_1\cdots i_n}=\sum_\mu
  \xi_{i_1}^\mu\cdots\xi_{i_n}^\mu.
  \label{eq:manybody}
\end{equation}
A literal $n$-body synapse is neither a standard hardware connection nor a
conventional biological synapse. Krotov and Hopfield's later ``large associative
memory'' formulation introduced feature neurons and memory neurons connected by
pairwise weights; eliminating the hidden population generates an effective
nonquadratic interaction among visible units.\cite{KrotovHopfield2021} This is a
familiar move in physics: an effective many-body term can arise after hidden
degrees of freedom are integrated out.

The distinction between \emph{parameter count} and \emph{state-space capacity}
is important here. Naively storing each $\bm{\xi}^\mu$ explicitly makes the
number of parameters grow with $K$. More recent random-feature formulations can
approximate dense associative energies with a fixed-size parameterization over
a chosen range.\cite{Hoover2024} An exponential number of stable states does not
mean an arbitrary exponential data set has been losslessly written into only
order-$N^2$ unconstrained real numbers.

\section{Exponential interactions and modern continuous Hopfield networks}
\label{sec:modern}

\subsection{From high powers to exponential capacity}

The polynomial result suggests making $F$ still sharper. Demircigil and
colleagues analyzed an exponential interaction and proved exponential storage
capacity under specified random-pattern, separation, and basin conditions.
\cite{Demircigil2017} ``Exponential capacity'' means
\begin{equation}
  K\sim e^{cN}
\end{equation}
with the explicit fixed-point condition $0<c<\ln 2/2$ for independent uniform
binary patterns; correction from a Hamming sphere imposes the stronger
basin-dependent bound in their theorem. It does not mean that
every one of the $2^N$ binary states can be stored with large, disjoint basins,
nor that correlations are irrelevant.

The mathematical mechanism is extreme selectivity. If the target overlap is
order $N$ and non-target overlaps are order $\sqrt{N}$, then
$\exp(\beta N)$ can dominate a sum of exponentially many
$\exp(\order{\beta\sqrt{N}})$ contributions until the entropy of competitors
overcomes the overlap gap. Capacity is an energy--entropy competition.

\subsection{A log-sum-exp energy}

Let $K$ continuous memory vectors $\bm{\xi}^\mu\in\mathbb{R}^d$ be columns of
\begin{equation}
  \bm{X}=[\bm{\xi}^1,\ldots,\bm{\xi}^K]\in\mathbb{R}^{d\times K}.
\end{equation}
A modern continuous Hopfield energy can be written, up to constants, as
\begin{equation}
  E(\bm{x})=\frac{1}{2}\lVert\bm{x}\rVert^2
  -\frac{1}{\beta}\log\sum_{\mu=1}^{K}
  \exp\!\left(\beta\bm{\xi}^{\mu\mathsf T}\bm{x}\right).
  \label{eq:modernenergy}
\end{equation}
Its gradient is
\begin{equation}
  \nabla E(\bm{x})=\bm{x}-
  \bm{X}\,\softmax(\beta\bm{X}^{\mathsf T}\bm{x}).
  \label{eq:moderngradient}
\end{equation}
The fixed-point update is therefore
\begin{equation}
  \boxed{
  \bm{x}_{t+1}=\bm{X}\,
  \softmax(\beta\bm{X}^{\mathsf T}\bm{x}_t)}.
  \label{eq:modernupdate}
\end{equation}
The softmax coefficients are nonnegative and sum to one, so each update is a
temperature-controlled convex combination of stored vectors. Large $\beta$
selects the strongest overlap; small $\beta$ averages more broadly. This makes
the competition between individual-memory fixed points and averaging fixed
points visible directly in the algebra.
The log-sum-exp term is convex, so the energy is a difference of convex
functions. The update can be derived by a convex--concave procedure that
majorizes the concave part, giving energy descent under the stated formulation.
The fixed points can include a global averaging state, metastable averages of a
subset, and states concentrated around individual patterns.
\cite{Ramsauer2021}

\begin{figure*}[t]
  \centering
  \includegraphics[width=0.94\textwidth]{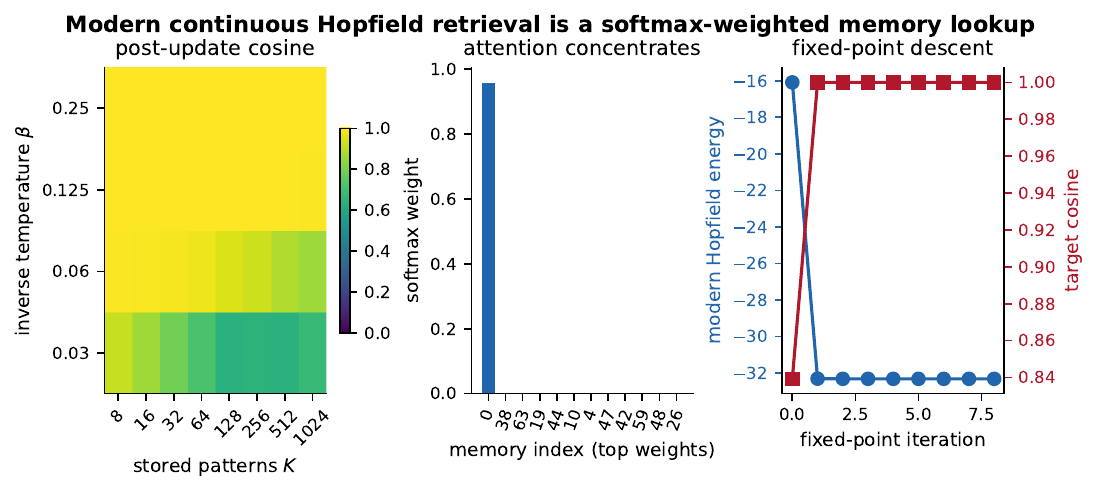}
  \caption{Modern continuous retrieval in $d=64$. Left: mean cosine similarity
  to a noisy target after one update, as the number of random stored patterns and
  inverse temperature vary. Middle: at adequate $\beta$, the softmax weights
  concentrate on the target memory. Right: repeated fixed-point updates lower
  Eq.~\eqref{eq:modernenergy} while increasing target similarity. These
  finite-dimensional simulations illustrate selectivity; they do not by
  themselves prove exponential capacity. Source:
  \texttt{code/hopfield\_replication.py}.}
  \label{fig:modern}
\end{figure*}

\subsection{The exact attention relation}

For a single query row vector $\bm{q}^{\mathsf T}$ and keys stored as the rows of
$\bm{K}$, scaled dot-product attention is
\begin{equation}
  \mathrm{Attn}(\bm{q},\bm{K},\bm{V})
  =\softmax\!\left(\frac{\bm{q}^{\mathsf T}\bm{K}^{\mathsf T}}
  {\sqrt{d_k}}\right)\bm{V}.
  \label{eq:attention}
\end{equation}
Set $\bm{V}=\bm{K}=\bm{X}^{\mathsf T}$ and
$\beta=1/\sqrt{d_k}$. After transposing conventions,
Eq.~\eqref{eq:attention} is exactly the update
Eq.~\eqref{eq:modernupdate}.\cite{Vaswani2017,Ramsauer2021,Widrich2020}

The equivalence is algebraically exact for this update, but its scope must be
stated.

\begin{itemize}[leftmargin=*]
  \item A transformer generally uses distinct learned query, key, and value
  projections. When values differ from keys, the output remains a softmax
  memory lookup but is not automatically the gradient fixed point of
  Eq.~\eqref{eq:modernenergy}.
  \item Multihead attention, residual streams, normalization, feedforward
  blocks, causal masking, and positional representations add dynamics absent
  from the elementary Hopfield system.
  \item Standard transformer inference applies a finite stack of different
  layers; it does not usually iterate one symmetric energy to equilibrium.
\end{itemize}

Thus ``attention is a modern Hopfield update'' can be a precise statement;
``a transformer is the 1982 Hopfield network'' is not.

\section{What, exactly, was the scientific advance?}
\label{sec:significance}

Hopfield's contribution was not the first associative memory and not the last
word on its capacity. Its significance can be stated in six parts.

\begin{enumerate}[leftmargin=*]
  \item \emph{Memory became a phase-space object.} Recall is convergence from a
  cue, not an address lookup. Basin geometry supplies a physical definition of
  error correction.

  \item \emph{A local rule received a global guarantee.} Symmetric interactions
  make each asynchronous decision descend the same scalar function. The proof
  scales to a large network without enumerating states.

  \item \emph{Robustness became collective.} A memory is distributed over the
  couplings and can survive component errors because the signal is reconstructed
  by many consistent contributions.

  \item \emph{Disorder became computational.} Other memories are not a small
  nuisance. Their crosstalk creates a capacity transition, false minima, and a
  spin-glass phase.

  \item \emph{Neural computation became accessible to statistical mechanics.}
  The model supplied order parameters and an ensemble in which typical behavior
  could be solved by mean-field and replica methods.

  \item \emph{The energy program remained extensible.} Graded units,
  optimization landscapes, high-order memories, hidden populations, exponential
  interactions, and softmax retrieval all preserve part of the original logic
  while changing other assumptions.
\end{enumerate}

\begin{table*}[t]
\caption{Precision strengthens the main claims.}
\label{tab:precision}
\begin{ruledtabular}
\begin{tabular}{p{0.28\textwidth}p{0.65\textwidth}}
Tempting shorthand & Precise statement \\
\hline
``Hopfield invented associative memory.'' & Associative and recurrent memory models predate 1982. Hopfield unified content-addressable recall, symmetric many-body dynamics, a Lyapunov energy, and a physical account of collective computation.\\
``Hebbian patterns are guaranteed minima.'' & At low load, random patterns are stable with high probability. Crosstalk, correlations, thresholds, finite size, and the diagonal convention can destabilize bits; spurious minima also occur.\\
``The Hopfield capacity is $0.14N$.'' & $\alpha_c\simeq0.138$ is the $T=0$ retrieval spinodal of the dense symmetric Hebbian model for unbiased random patterns at $N\to\infty$. Other criteria and learning rules give other numbers.\\
``Energy descent solves optimization.'' & Descent proves convergence to a fixed point or stationary point under the stated dynamics. It does not prove arrival at the global optimum or a desired memory.\\
``Dense memory stores $N^{n-1}$ patterns.'' & Polynomial order $n$ permits fixed-bit-error capacity of that order under random-pattern assumptions. Whole-pattern error control adds a $\ln N$ correction and constants depend on conventions.\\
``Modern Hopfield networks store exponentially many arbitrary memories.'' & Exponential capacity requires norm and separation conditions and a specified retrieval error/basin criterion; correlated or insufficiently separated patterns can merge.\\
``Attention is a Hopfield network.'' & A special modern Hopfield fixed-point update equals scaled dot-product attention when keys and values are the stored patterns. A full transformer has additional learned, asymmetric, depth-dependent operations.\\
\end{tabular}
\end{ruledtabular}
\end{table*}

\section{Limits, biological interpretation, and effective scale}
\label{sec:limits}

The classical model is often evaluated with the wrong standard. It is too
idealized to be a cellular reconstruction, yet too mathematically specific to
serve as an unconstrained metaphor. Its proper role is an effective theory: it
identifies sufficient circuit-level conditions for attractor memory and exposes
the variables---overlap, load, susceptibility, temperature, and basin
geometry---that control that regime.

Symmetric coupling is especially consequential. Real chemical synapses are
directed, and generic recurrent circuits need not possess an energy. Approximate
symmetry can still produce slow attractor-like modes, but delays, adaptation,
spike timing, ongoing input, and asymmetry can create cycles, sequential
activity, metastability, or chaos. Those behaviors are not ``violations'' of
the Hopfield proof; they lie outside its assumptions.

The Hebbian rule Eq.~\eqref{eq:hebb} is also an embedding prescription rather
than a complete biological learning theory. It assumes access to the patterns,
adds their outer products, and often centers them. Synaptic locality alone does
not settle how patterns are presented, how homeostasis prevents runaway
coupling, how interference is managed over time, or how memories consolidate.

Finally, point-neuron convenience does not guarantee scale sufficiency. A
single cell may contain dendritic subunits with nonlinear local integration;
neuromodulation can change effective gain and plasticity; and circuit morphology
can constrain which symmetric or low-rank interactions are realizable. Such
details may be integrated out when they only renormalize $g$, $J$, or noise. If
they create additional state variables or path-dependent interactions on the
recall timescale, the effective Hopfield description must be enlarged. The
right scale is determined by closure of the macroscopic dynamics, not by ease of
simulation.

This caution does not diminish the model. On the contrary, it gives a concrete
experimental program: measure whether perturbations relax toward reproducible
population states; estimate basin geometry; test whether an energy-like scalar
decreases; and determine which omitted variables predict deviations from the
reduced dynamics.

\section{Computational companion and reproducibility}
\label{sec:repro}

The supplied script uses NumPy, SciPy, and Matplotlib and requires no external
data. From the project directory, run
\begin{verbatim}
python code/hopfield_replication.py \
  --output figures \
  --results numerical_results.json
\end{verbatim}
The fixed seed is $1982$ unless overridden. The \texttt{--quick} flag reduces
Monte Carlo repetitions. Each figure is saved as PDF, PNG, and SVG. The JSON
file records model sizes, loads, overlaps, energy changes, capacity curves, and
modern-retrieval cosines.

\begin{table}[b]
\caption{What each numerical figure establishes.}
\label{tab:figures}
\begin{ruledtabular}
\begin{tabular}{p{0.19\columnwidth}p{0.72\columnwidth}}
Figure & Reproduced mechanism \\
\hline
\ref{fig:classical} & asynchronous energy descent and completion from a corrupted cue \\
\ref{fig:capacity} & $T=0$ mean-field continuation and a distinct finite-$N$ retrieval assay \\
\ref{fig:graded} & continuous-state Lyapunov descent for monotone graded units \\
\ref{fig:dense} & change of polynomial capacity exponent and exact bit-stability test \\
\ref{fig:modern} & softmax concentration, fixed-point descent, and target retrieval \\
\end{tabular}
\end{ruledtabular}
\end{table}

Numerical evidence should be read at the level it supports. Figure
\ref{fig:classical} does not measure capacity. Figure \ref{fig:capacity}'s
finite-$N$ dynamics is not the replica calculation. Figure \ref{fig:dense}
checks stability at an asymptotic scale but does not fit the exponent from many
decades of $N$. Figure \ref{fig:modern} demonstrates selective retrieval but
does not prove exponential capacity. The analytic arguments supply those wider
claims; the code makes their mechanisms inspectable.

\section{Derivations and exercises}
\label{sec:exercises}

\subsection{Binary dynamics}

\begin{enumerate}[leftmargin=*]
  \item Re-derive Eq.~\eqref{eq:deltaenergygeneral} without assuming symmetry.
  Show explicitly which unmatched term remains when $J_{ij}\ne J_{ji}$.
  \item For two synchronous spins with symmetric negative coupling, construct a
  two-cycle. Explain why this does not contradict asynchronous descent.
  \item Retain nonzero thresholds and convert between $V_i\in\{0,1\}$ and
  $s_i\in\{-1,+1\}$. Track the constant and linear terms in the energy.
  \item Modify the script so updates always follow the same index order. Compare
  convergence time and final attractor with random sequential order.
\end{enumerate}

\subsection{Crosstalk and phase structure}

\begin{enumerate}[leftmargin=*]
  \item Starting at $\bm{s}=\bm{\xi}^1$, derive
  Eq.~\eqref{eq:naivevariance} including the $J_{ii}=0$ correction. Which terms
  vanish only after averaging over patterns?
  \item Compute the three-pattern mixture overlap exactly by enumerating the
  eight possible triples $(\xi_i^1,\xi_i^2,\xi_i^3)$.
  \item Solve Eqs.~\eqref{eq:t0m}--\eqref{eq:t0r} independently at each
  $\alpha$ with initial guess $m\approx0$, then by continuation from small
  $\alpha$. Explain why the first procedure can miss the retrieval branch.
  \item At finite temperature, numerically solve
  Eqs.~\eqref{eq:rsm}--\eqref{eq:rsr}. Plot $m$, $q$, and $C$ separately. Do not
  label a regime from $m$ alone.
\end{enumerate}

\subsection{Graded and optimization models}

\begin{enumerate}[leftmargin=*]
  \item Differentiate Eq.~\eqref{eq:tanhintegral} to recover
  $g^{-1}(V)=\gamma^{-1}\mathop{\mathrm{atanh}}V$.
  \item Replace $T$ in Eq.~\eqref{eq:gradedode} by $T+A$, where $A$ is
  antisymmetric. Evaluate $\dd E/\dd t$ using the symmetric-part energy and
  identify why its sign is no longer fixed.
  \item Encode a small assignment problem with penalty terms. Enumerate all
  binary states, compare their true cost with network energy, and locate invalid
  low-energy states as the penalty strength varies.
\end{enumerate}

\subsection{Dense memory and attention}

\begin{enumerate}[leftmargin=*]
  \item Derive Eqs.~\eqref{eq:densesignal} and \eqref{eq:densevariance} for
  $n=3$ without using the general double-factorial formula.
  \item Explain why fixed bit-error probability gives
  Eq.~\eqref{eq:densefixed}, whereas controlling any error among $N$ bits
  introduces $\ln N$ in Eq.~\eqref{eq:densenone}.
  \item Differentiate Eq.~\eqref{eq:modernenergy} and recover
  Eq.~\eqref{eq:modernupdate}. Then use distinct value vectors and identify the
  step at which the strict gradient-energy interpretation is lost.
  \item Construct a small correlated family analytically (for example, memories
  sharing a common component) and predict which pair will merge first under
  Eq.~\eqref{eq:modernupdate}. Optionally verify the prediction by modifying
  Fig.~\ref{fig:modern}. Relate retrieval failure to pairwise separation rather
  than to $K$ alone.
\end{enumerate}

\section{Primary literature and further reading}
\label{sec:reading}

Read the two PNAS papers in sequence. In Hopfield 1982, mark separately the
phase-space definition of content addressability, the coupling prescription,
the energy argument, and the simulations.\cite{Hopfield1982} In Hopfield 1984,
derive the inverse-response integral and ask exactly why monotonicity and
symmetry are sufficient.\cite{Hopfield1984}

Then read Little and at least one of Amari, Anderson, or Kohonen before making a
priority claim.\cite{Little1974,Amari1972,Anderson1972,Kohonen1972} This places
the Hopfield synthesis in its actual lineage. Follow with the 1985 PRL and the
1987 Annals of Physics paper by Amit, Gutfreund, and Sompolinsky; the former
announces extensive storage, while the latter provides the fuller near-saturation
theory.\cite{AmitGutfreundSompolinsky1985PRL,AmitGutfreundSompolinsky1987}

To understand the style of reasoning behind this trajectory, pair Hopfield's
reflection on working across physics and biology with his Nobel lecture, then
read Bialek's prize-era account of emergence and the expanding boundaries of
physics.\cite{Hopfield2014TwoCultures,Hopfield2025NobelLecture,Bialek2025Emergence}

For the later arc, read Krotov and Hopfield 2016 for polynomial energies,
energy-difference updates, capacity scaling, and the feature--prototype
duality.\cite{KrotovHopfield2016} Pair it with Demircigil et al. for the
exponential-capacity theorem, then Krotov and Hopfield 2021 for a pairwise
visible--hidden implementation.\cite{Demircigil2017,KrotovHopfield2021} Only
afterward read Ramsauer et al. alongside the original attention paper.
\cite{Ramsauer2021,Vaswani2017} That order makes the softmax relation a derived
continuation of an energy program rather than a retrospective slogan.

\section{What survives from the Hopfield program?}

\paragraph{Established result.}
Hopfield's classical construction proves that symmetric couplings and
asynchronous local updates can make a recurrent network descend a global
Lyapunov function. With Hebbian superposition, designated patterns become
content-addressable attractors at low load. For independent unbiased random
patterns at extensive load, the Amit--Gutfreund--Sompolinsky theory identifies
retrieval and spin-glass order parameters and locates the zero-temperature
retrieval spinodal near $\alpha_c\simeq0.138$. These statements are precise
because their state variables, dynamics, disorder ensemble, thermodynamic
limit, and success criterion are explicit.

\paragraph{Surviving principle.}
The enduring idea is not that cognition is energy minimization. It is that a
computation can be defined at the level of a collective dynamical state and
analyzed with the tools of physics. In this formulation, a cue is an initial
condition, recall is a trajectory, error correction is convergence within a
basin, interference is crosstalk from quenched disorder, and failure is a
reorganization of the macroscopic landscape. Graded units, polynomial and
exponential memories, hidden-population implementations, and log-sum-exp
retrieval retain this logic while altering the interaction, state space, and
capacity scaling.

\paragraph{Limitation.}
Energy descent guarantees neither the desired memory nor a global optimum, and
the classical assumptions exclude generic directed, delayed, adaptive, and
driven recurrent dynamics. The Hebbian outer-product rule is an embedding
prescription rather than a complete theory of biological learning. Likewise,
the exact equivalence between a modern Hopfield update and scaled dot-product
attention holds only under a particular identification of queries, keys, and
values; it does not make a full transformer an equilibrium Hopfield network.
These limitations specify the boundary of the effective theory rather than
reducing it to a metaphor.

\paragraph{Open problem.}
The central problem for theoretical NeuroAI is to determine which parts of the
attractor description remain closed under biologically and computationally
relevant extensions. How much asymmetry, ongoing drive, synaptic dynamics,
dendritic state, or structured correlation can be integrated out into effective
couplings and noise before memory requires a genuinely nonequilibrium theory?
Answering this demands a common language for basin geometry, transient
computation, sequence generation, capacity, and learnability across biological
and artificial systems. The methodological standard is therefore to state the
degrees of freedom, dynamics, symmetry, ensemble, scaling limit, order
parameters, and retrieval criterion---and then test which omitted variables
change the regime.

\section*{Data $\&$ Code Availability}

Code and figure-generation script are available at: 
\url{https://github.com/neurovium/NeuroAI}.

\section*{References}

% Chapter-scoped database retained separately for later book-wide merging.
\bibliography{hopfield_book_chapter_references}

@article{McCullochPitts1943,
  author = {McCulloch, Warren S. and Pitts, Walter},
  title = {A Logical Calculus of the Ideas Immanent in Nervous Activity},
  journal = {Bulletin of Mathematical Biophysics},
  year = {1943},
  volume = {5},
  pages = {115--133},
  doi = {10.1007/BF02478259}
}

@book{Hebb1949,
  author = {Hebb, Donald O.},
  title = {The Organization of Behavior: A Neuropsychological Theory},
  publisher = {Wiley},
  address = {New York},
  year = {1949}
}

@article{Amari1972,
  author = {Amari, Shun-ichi},
  title = {Learning Patterns and Pattern Sequences by Self-Organizing Nets of Threshold Elements},
  journal = {IEEE Transactions on Computers},
  year = {1972},
  volume = {C-21},
  number = {11},
  pages = {1197--1206},
  doi = {10.1109/T-C.1972.223477}
}

@article{Anderson1972,
  author = {Anderson, James A.},
  title = {A Simple Neural Network Generating an Interactive Memory},
  journal = {Mathematical Biosciences},
  year = {1972},
  volume = {14},
  pages = {197--220},
  doi = {10.1016/0025-5564(72)90075-2}
}

@article{Kohonen1972,
  author = {Kohonen, Teuvo},
  title = {Correlation Matrix Memories},
  journal = {IEEE Transactions on Computers},
  year = {1972},
  volume = {C-21},
  number = {4},
  pages = {353--359},
  doi = {10.1109/TC.1972.5008975}
}

@article{Little1974,
  author = {Little, William A.},
  title = {The Existence of Persistent States in the Brain},
  journal = {Mathematical Biosciences},
  year = {1974},
  volume = {19},
  pages = {101--120},
  doi = {10.1016/0025-5564(74)90031-5}
}

@article{Hopfield1958,
  author = {Hopfield, John J.},
  title = {Theory of the Contribution of Excitons to the Complex Dielectric Constant of Crystals},
  journal = {Physical Review},
  year = {1958},
  volume = {112},
  number = {5},
  pages = {1555--1567},
  doi = {10.1103/PhysRev.112.1555}
}

@article{HopfieldElectronTransfer1974,
  author = {Hopfield, John J.},
  title = {Electron Transfer Between Biological Molecules by Thermally Activated Tunneling},
  journal = {Proceedings of the National Academy of Sciences USA},
  year = {1974},
  volume = {71},
  number = {9},
  pages = {3640--3644},
  doi = {10.1073/pnas.71.9.3640}
}

@article{HopfieldProofreading1974,
  author = {Hopfield, John J.},
  title = {Kinetic Proofreading: A New Mechanism for Reducing Errors in Biosynthetic Processes Requiring High Specificity},
  journal = {Proceedings of the National Academy of Sciences USA},
  year = {1974},
  volume = {71},
  number = {10},
  pages = {4135--4139},
  doi = {10.1073/pnas.71.10.4135}
}

@article{Hopfield2014TwoCultures,
  author = {Hopfield, John J.},
  title = {Two Cultures? Experiences at the Physics--Biology Interface},
  journal = {Physical Biology},
  year = {2014},
  volume = {11},
  number = {5},
  pages = {053002},
  doi = {10.1088/1478-3975/11/5/053002}
}

@article{Hopfield2025NobelLecture,
  author = {Hopfield, John J.},
  title = {Nobel Lecture: Physics Is a Point of View},
  journal = {Reviews of Modern Physics},
  year = {2025},
  volume = {97},
  number = {3},
  pages = {030501},
  doi = {10.1103/RevModPhys.97.030501}
}

@article{Bialek2025Emergence,
  author = {Bialek, William},
  title = {Emergence of Brains},
  journal = {PRX Life},
  year = {2025},
  volume = {3},
  number = {3},
  pages = {037002},
  doi = {10.1103/62p9-wn32}
}

@article{Hopfield1982,
  author = {Hopfield, John J.},
  title = {Neural Networks and Physical Systems with Emergent Collective Computational Abilities},
  journal = {Proceedings of the National Academy of Sciences USA},
  year = {1982},
  volume = {79},
  number = {8},
  pages = {2554--2558},
  doi = {10.1073/pnas.79.8.2554}
}

@article{Hopfield1984,
  author = {Hopfield, John J.},
  title = {Neurons with Graded Response Have Collective Computational Properties Like Those of Two-State Neurons},
  journal = {Proceedings of the National Academy of Sciences USA},
  year = {1984},
  volume = {81},
  number = {10},
  pages = {3088--3092},
  doi = {10.1073/pnas.81.10.3088}
}

@article{CohenGrossberg1983,
  author = {Cohen, Michael A. and Grossberg, Stephen},
  title = {Absolute Stability of Global Pattern Formation and Parallel Memory Storage by Competitive Neural Networks},
  journal = {IEEE Transactions on Systems, Man, and Cybernetics},
  year = {1983},
  volume = {SMC-13},
  number = {5},
  pages = {815--826},
  doi = {10.1109/TSMC.1983.6313075}
}

@article{HopfieldTank1985,
  author = {Hopfield, John J. and Tank, David W.},
  title = {``Neural'' Computation of Decisions in Optimization Problems},
  journal = {Biological Cybernetics},
  year = {1985},
  volume = {52},
  pages = {141--152},
  doi = {10.1007/BF00339943}
}

@article{AmitGutfreundSompolinsky1985PRA,
  author = {Amit, Daniel J. and Gutfreund, Hanoch and Sompolinsky, Haim},
  title = {Spin-Glass Models of Neural Networks},
  journal = {Physical Review A},
  year = {1985},
  volume = {32},
  number = {2},
  pages = {1007--1018},
  doi = {10.1103/PhysRevA.32.1007}
}

@article{AmitGutfreundSompolinsky1985PRL,
  author = {Amit, Daniel J. and Gutfreund, Hanoch and Sompolinsky, Haim},
  title = {Storing Infinite Numbers of Patterns in a Spin-Glass Model of Neural Networks},
  journal = {Physical Review Letters},
  year = {1985},
  volume = {55},
  number = {14},
  pages = {1530--1533},
  doi = {10.1103/PhysRevLett.55.1530}
}

@article{AmitGutfreundSompolinsky1987,
  author = {Amit, Daniel J. and Gutfreund, Hanoch and Sompolinsky, Haim},
  title = {Statistical Mechanics of Neural Networks Near Saturation},
  journal = {Annals of Physics},
  year = {1987},
  volume = {173},
  number = {1},
  pages = {30--67},
  doi = {10.1016/0003-4916(87)90092-3}
}

@article{SteffanKuhn1994,
  author = {Steffan, Helmut and K{\"u}hn, Reimer},
  title = {Replica Symmetry Breaking in Attractor Neural Network Models},
  journal = {Zeitschrift f{\"u}r Physik B: Condensed Matter},
  year = {1994},
  volume = {95},
  number = {2},
  pages = {249--260},
  doi = {10.1007/BF01312198}
}

@article{McEliece1987,
  author = {McEliece, Robert J. and Posner, Edward C. and Rodemich, Eugene R. and Venkatesh, Santosh S.},
  title = {The Capacity of the {Hopfield} Associative Memory},
  journal = {IEEE Transactions on Information Theory},
  year = {1987},
  volume = {33},
  number = {4},
  pages = {461--482},
  doi = {10.1109/TIT.1987.1057328}
}

@article{Personnaz1986,
  author = {Personnaz, L. and Guyon, I. and Dreyfus, G.},
  title = {Collective Computational Properties of Neural Networks: New Learning Mechanisms},
  journal = {Physical Review A},
  year = {1986},
  volume = {34},
  number = {5},
  pages = {4217--4228},
  doi = {10.1103/PhysRevA.34.4217}
}

@article{Gardner1988,
  author = {Gardner, Elizabeth},
  title = {The Space of Interactions in Neural Network Models},
  journal = {Journal of Physics A: Mathematical and General},
  year = {1988},
  volume = {21},
  number = {1},
  pages = {257--270},
  doi = {10.1088/0305-4470/21/1/030}
}

@article{TsodyksFeigelman1988,
  author = {Tsodyks, Mikhail V. and Feigel'man, Mikhail V.},
  title = {The Enhanced Storage Capacity in Neural Networks with Low Activity Level},
  journal = {Europhysics Letters},
  year = {1988},
  volume = {6},
  number = {2},
  pages = {101--105},
  doi = {10.1209/0295-5075/6/2/002}
}

@book{Amit1989,
  author = {Amit, Daniel J.},
  title = {Modeling Brain Function: The World of Attractor Neural Networks},
  publisher = {Cambridge University Press},
  address = {Cambridge},
  year = {1989},
  doi = {10.1017/CBO9780511623257}
}

@article{Kanter1988,
  author = {Kanter, Ido},
  title = {Potts-Glass Models of Neural Networks},
  journal = {Physical Review A},
  year = {1988},
  volume = {37},
  number = {7},
  pages = {2739--2742},
  doi = {10.1103/PhysRevA.37.2739}
}

@inproceedings{KrotovHopfield2016,
    author = {Krotov, Dmitry and Hopfield, John J.},
    title = {Dense associative memory for pattern recognition},
    year = {2016},
    isbn = {9781510838819},
    publisher = {Curran Associates Inc.},
    address = {Red Hook, NY, USA},
    booktitle = {Proceedings of the 30th International Conference on Neural Information Processing Systems},
    pages = {1180–1188},
    numpages = {9},
    location = {Barcelona, Spain},
    series = {NIPS'16}
}

@article{KrotovHopfield2018,
  author = {Krotov, Dmitry and Hopfield, John J.},
  title = {Dense Associative Memory Is Robust to Adversarial Inputs},
  journal = {Neural Computation},
  year = {2018},
  volume = {30},
  number = {12},
  pages = {3151--3167},
  doi = {10.1162/neco_a_01143}
}

@article{Demircigil2017,
  author = {Demircigil, Mete and Heusel, Judith and L\"owe, Matthias and Upgang, Sven and Vermet, Franck},
  title = {On a Model of Associative Memory with Huge Storage Capacity},
  journal = {Journal of Statistical Physics},
  year = {2017},
  volume = {168},
  number = {2},
  pages = {288--299},
  doi = {10.1007/s10955-017-1806-y}
}

@inproceedings{KrotovHopfield2021,
  author = {Krotov, Dmitry and Hopfield, John J.},
  title = {Large Associative Memory Problem in Neurobiology and Machine Learning},
  booktitle = {International Conference on Learning Representations},
  year = {2021},
  url = {https://openreview.net/forum?id=X4y_10OX-hX}
}

@inproceedings{Ramsauer2021,
  author = {Ramsauer, Hubert and Sch\"afl, Bernhard and Lehner, Johannes and Seidl, Philipp and Widrich, Michael and Adler, Thomas and Gruber, Lukas and Holzleitner, Markus and Pavlovi\'c, Milena and Sandve, Geir Kjetil and Greiff, Victor and Kreil, David and Kopp, Michael and Klambauer, G\"unter and Brandstetter, Johannes and Hochreiter, Sepp},
  title = {Hopfield Networks Is All You Need},
  booktitle = {International Conference on Learning Representations},
  year = {2021},
  url = {https://arxiv.org/abs/2008.02217}
}

@inproceedings{Vaswani2017,
 author = {Vaswani, Ashish and Shazeer, Noam and Parmar, Niki and Uszkoreit, Jakob and Jones, Llion and Gomez, Aidan N and Kaiser, \L ukasz and Polosukhin, Illia},
 booktitle = {Advances in Neural Information Processing Systems},
 editor = {I. Guyon and U. Von Luxburg and S. Bengio and H. Wallach and R. Fergus and S. Vishwanathan and R. Garnett},
 pages = {},
 publisher = {Curran Associates, Inc.},
 title = {Attention is All you Need},
 url = {https://proceedings.neurips.cc/paper_files/paper/2017/file/3f5ee243547dee91fbd053c1c4a845aa-Paper.pdf},
 url = {https://arxiv.org/abs/1706.03762},
 volume = {30},
 year = {2017}
}

@inproceedings{Widrich2020,
    author = {Widrich, Michael and Sch{\"a}fl, Bernhard and Pavlovi{\'c}, Milena and Ramsauer, Hubert and Gruber, Lukas and Holzleitner, Markus and Brandstetter, Johannes and Sandve, Geir Kjetil and Greiff, Victor and Hochreiter, Sepp and Klambauer, G{\"u}nter},
    title = {Modern hopfield networks and attention for immune repertoire classification},
    year = {2020},
    isbn = {9781713829546},
    publisher = {Curran Associates Inc.},
    address = {Red Hook, NY, USA},
    booktitle = {Proceedings of the 34th International Conference on Neural Information Processing Systems},
    articleno = {1581},
    numpages = {14},
    location = {Vancouver, BC, Canada},
    series = {NIPS '20}
}

@inproceedings{Hoover2024,
    author = {Hoover, Benjamin and Chau, Duen Horng and Strobelt, Hendrik and Ram, Parikshit and Krotov, Dmitry},
    title = {Dense associative memory through the lens of random features},
    year = {2024},
    isbn = {9798331314385},
    publisher = {Curran Associates Inc.},
    address = {Red Hook, NY, USA},
    booktitle = {Proceedings of the 38th International Conference on Neural Information Processing Systems},
    articleno = {742},
    numpages = {28},
    location = {Vancouver, BC, Canada},
    series = {NIPS '24}
}

@misc{Nobel2024,
  author = {{Royal Swedish Academy of Sciences}},
  title = {Scientific Background to the {Nobel Prize in Physics 2024}: Machine Learning with Artificial Neural Networks},
  year = {2024},
  url = {https://www.nobelprize.org/prizes/physics/2024/advanced-information/}
}

\end{document}